\documentclass[graybox]{svmult}

\usepackage[l2tabu,orthodox]{nag}

\usepackage[sort, square,numbers]{natbib}
\usepackage[ pdftex,colorlinks]{hyperref} %

\usepackage{layouts} %

\usepackage[printonlyused]{acronym}
\usepackage{siunitx}
\usepackage[all]{nowidow}

\usepackage[dvipsnames, table]{xcolor}

\usepackage{lipsum}

\usepackage{verbatim}

\usepackage{pdflscape}

\usepackage{pifont}
\usepackage[pdftex]{graphicx}
\usepackage{epstopdf}

\usepackage{import}

\usepackage{float}

\usepackage[font=footnotesize]{caption}

\graphicspath{{./latexGoodPractices/}}

\usepackage{booktabs}

\usepackage{tabu}

\usepackage{supertabular}

\usepackage{rotating,tabularx}

\usepackage{makecell}

\usepackage{amssymb,amsfonts,amsmath,amscd}

\usepackage{bm}

\newcommand{\bbm}{\begin{bmatrix}}
\newcommand{\ebm}{\end{bmatrix}}

\acrodef{ICP}{Iterative Closest Point}
\acrodef{LIDAR}{Light Detection And Ranging}
\acrodef{L2}{squared distance}
\acrodef{IRLS}{Iteratively Reweighted Least-Squares}
\acrodef{SLAM}{Simultaneous Localization and Mapping}
\acrodef{DBH}{diameter at breast height}
\acrodef{TLS}{terrestrial laser scanning}
\acrodef{DTM}{digital terrain model}
\acrodef{RMSE}{root mean square error}
\acrodef{IMU}{inertial measurement unit}
\acrodef{EKF}{extended Kalman filter}

\usepackage{algorithm}
\usepackage{algpseudocode}
\usepackage{mathptmx}       %
\usepackage{helvet}         %
\usepackage{courier}        %
\usepackage{type1cm}        %
\usepackage{makeidx}         %
\usepackage{graphicx}        %
\usepackage{multicol}        %
\usepackage[bottom]{footmisc}%
\usepackage{subcaption}
\usepackage[inline]{enumitem}
\usepackage[percent]{overpic}

\usepackage{hhline}

\usepackage{sidecap}
\sidecaptionvpos{figure}{c}

\usepackage[normalem]{ulem}  %

\usepackage{tikz}
\usetikzlibrary{positioning}
\usetikzlibrary{arrows.meta}
\tikzset{%
  >={Latex[width=2mm,length=2mm]},
            base/.style = {rectangle, rounded corners, draw=black,
                           minimum width=4cm, minimum height=1cm,
                           text centered, font=\rmfamily},
         normalBlock/.style = {base, minimum width=2.5cm, fill=orange!15,
                           font=\rmfamily},
         endBlock/.style = {base, minimum width=2.5cm, fill=green!15,
         font=\rmfamily},
         beginningBlock/.style = {base, minimum width=2.5cm, fill=red!15,
         font=\rmfamily},
}

\DeclareMathOperator*{\argmin}{arg\,min}

\makeindex

\begin{document}

\title*{Automatic 3D Mapping for Tree Diameter Measurements in Inventory Operations}
\author{Jean-Fran\c cois Tremblay$^{*}$, Martin B\'eland$^{+}$, Fran\c cois Pomerleau$^{*}$, Richard Gagnon$^{\dagger}$, Philippe Gigu\`ere\hspace{-3pt}
	\thanks{Northern Robotics Laboratory, Universit\'e Laval\newline
    + Department of Geomatics Sciences, Universit\'e Laval\newline
    $\dagger$ Centre de Recherche Industrielle du Qu\'ebec\newline
		Communication e-mail: {\tt\small jean-francois.tremblay.36@ulaval.ca }}}
\authorrunning{J.-F. Tremblay et al.}
\maketitle

\abstract{
  Forestry is a major industry in many parts of the world.
  It relies on forest inventory, which consists of measuring tree attributes.
  We propose to use 3D mapping, based on the iterative closest point algorithm, to automatically measure tree diameters in forests from mobile robot observations.
  While previous studies showed the potential for such technology, they lacked a rigorous analysis of diameter estimation methods in challenging forest environments.
  Here, we validated multiple diameter estimation methods, including two novel ones, in a new varied dataset of four different forest sites, 11 trajectories, totalling 1458 tree observations and 1.4 hectares.
  We provide recommendations for the deployment of mobile robots in a forestry context.
  We conclude that our mapping method is usable in the context of automated forest inventory, with our best method yielding a root mean square error of \SI{3.45}{\cm} for our whole dataset, and \SI{2.04}{\cm} in ideal conditions consisting of mature forest with well spaced trees.
}

\section{Introduction}
\label{sec:intro}

Forestry is an important industry in many countries.
In 2016, it accounted for about 13 billion USD in Canada's economy and a similar figure in Sweden's exports of wood products.
Yet, worker shortages and high turnover rates coupled with long training time are threatening many operations in this industry.
Recent progress in field robotics, such as 3D mapping, have the potential to improve forestry operations while reducing demand for labor.
Furthermore, these 3D mapping technologies could be used to estimate wood biomass for carbon accounting purposes~\citep{biomass}.
From a scientific point of view, studying the wider context of field robotics in forests is interesting, from the new challenges it generates.
For instance, localization and mapping is more difficult in unstructured environments~\citep{Pomerleau2013}.

\begin{figure}[h!]
  \includegraphics[width=\linewidth, trim={0 12.5cm 0 2cm}, clip]{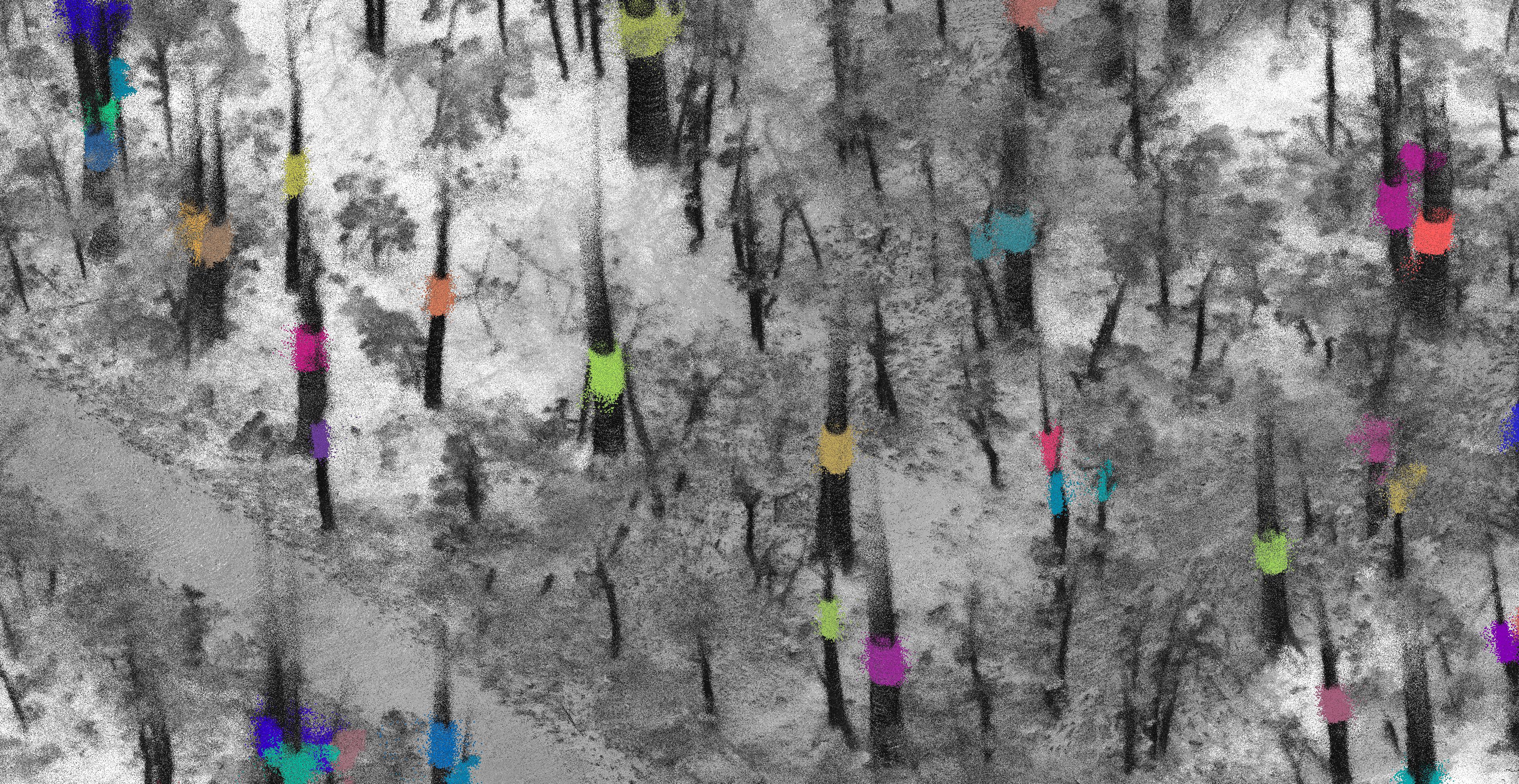}
  \caption{
  Perspective view of a forest continuously mapped by a ground vehicle equipped with a 3D lidar, using the method described in \autoref{sec:mapping}.
  Our focus is on how to best estimate the diameter at breast height, which is highlighted in an individual color for each tree in this figure.
  }
  \label{fig-OverviewMap}
\vspace{-15pt}
\end{figure}

A key component in modern forest operations is forest inventory~\citep{tlsinventory}.
It consists in identifying specific attributes of trees.
Some of these attributes, such as species, can be estimated using cameras and advanced computer vision techniques \citep{mathieu}.
Others, such as tree diameters, can be extracted from lidar point clouds, as they contain metric information.
We conjecture that the ability for an autonomous system to process geometric information via 3D mapping is one of the key elements to the development of future intelligent forest machinery.
In our immediate case of forest inventory, this would enable automatic or computer-assisted tree selection for forest harvesting equipment.
At the moment, deciding which trees to harvest in a partial cut scenario is performed manually by a technician.
This operation has been identified as expensive, time consuming as well as yielding different results depending on the technician~\citep{markingunreliable}.
We believe that this could be addressed by equipping harvesting machinery with the proper sensors and algorithms.
This paper explores the use of automatic map building with a standard set of robotic sensors, within the context of forest inventory.
Although full 3D maps are produced in the process, as shown in \autoref{fig-OverviewMap}, we limit our quantitative study on a single standard attribute in the forest inventory: \acf{DBH} measurements.
The \ac{DBH} is arguably the most important tree characteristic used for tree selection and wood volume prediction in the forest industry~\citep{west2009tree}.
Typical requirements for diameter measurement accuracy are around \SI{2}{\cm} of error \citep{tlsinventory}, but can be as high as \SI{5}{\cm} for American diameter classes \citep{americandbh}.
Early work on tree diameter estimation from lidars focused on \acf{TLS}~\citep{tlsinventory}.
The \ac{TLS}, which is mounted on a tripod, is manually moved by an operator.
Once the individual scans are registered using markers manually installed in the environment, one obtains a very precise 3D map of the environment.
However, this data collection approach is significantly more tedious and time consuming than mobile mapping techniques.

We propose to use an \ac{ICP} 3D mapping approach for forest mapping.
A map in 2D might not be sufficient for forest mapping or forest robotics in general. Indeed, forests are rarely flat and even when they are, obstacles on the ground will result on the robot not being levelled, as will be shown in our dataset.
Our dataset will present one particularly steep forest where it is not clear how any 2D approach could work.
Point clouds generated from this approach tend to be noisier than \ac{TLS} in nature \cite{tlsinventory}; this causes problems in the accuracy of diameter extraction.
From the generated 3D maps, we automatically estimate tree diameters, comparing several approaches.
We validate our complete approach on an extensive dataset of 11 trajectories through four forests, varying topographies, ages, species compositions and densities.

\renewcommand\labelenumi{(\theenumi)}
Our contributions are as follows:
\begin{enumerate*}
  \item we test ICP mapping in different forest types for the first time and provide insight about its performance and limitations;
  \item we propose a new robust approach to \acf{DBH} estimation, based on the median of several cylinder fittings, designed to perform well on noisy maps, including those built with ICP;
  \item we perform extensive validation of different tree diameter estimation methods from noisy lidar data, and identify which ones perform best; and
  \item we provide recommendations on trajectories and field deployment for diameter estimation from robot mapping.
\end{enumerate*}

\section{Related works}
\label{sec:relatedworks}
Before measuring tree \ac{DBH}s from lidar-equipped mobile platforms, one needs to create a map from the observations.
\citet{jagbrant2015lidar} used a 2D lidar, combined with GPS and a IMU only for localization, to detect trees in orchards.
This approach is not optimal in natural forests as GPS performance is affected by heavy forest canopy.
\citet{tsubouchi2014forest} bridged the gap to natural forests, using a 2D lidar on a pan-tilt unit combined with a tripod.
The scans were taken in a static manner as opposed to a moving robot.
The mapping was done using a combination of the tree's location and \ac{ICP}.
\citet{tang2015slam} created maps using \acf{SLAM}, for trajectories traveling on a road through the forest as opposed to inside the forest itself.
This trajectory was selected to improve GPS reception.
Importantly, they did not perform full 3D mapping which, as we claim, is key to enabling robotics in forests.
\citet{calderscomp} performed an extensive comparison between a commercial GeoSLAM handheld scanner and \ac{TLS}.
Their testing was limited to circular plots with \SI{15}{\m} radius.
They also noted that the \ac{SLAM} still failed on two of the forest plots tested.
More recently, \citet{seiki2017backpack} scanned a forest with a 2D lidar and a pan-tilt unit on a backpack, using a \ac{SLAM} technique called LOAM \citep{loam}.
Other work using graph-\ac{SLAM} has been done in~\cite{marek}.

After generating a 3D map, one needs to use either a circle fitting or cylinder fitting algorithm to estimate \ac{DBH}s.
\citet{jfrDetectionAndDTM} presented a method to detect and segment trees in static lidar scans.
They tested diameter estimation using cone and cylinder fitting for five sites.
They reported a \ac{RMSE} of more than \SI{13}{\cm}, which is not sufficiently accurate for forest inventory.
\citet{tsubouchi2014forest} performed least square 2D circle fitting for diameter estimation.
\citet{calderscomp} used Computree~\citep{computree} for terrain height models and diameter measurements, which was developed for \ac{TLS}.
\citet{seiki2017backpack} employed ``the Point Cloud Library RANSAC cylinder fitting method''.
In \citep{marek}, two circle fitting methods were validated: the Pratt fit \citep{pratt1987direct} and least square circle fitting.

Another important aspect is the rigorous analysis of the mapping and diameter extraction method, both in terms of forest variety and the number of trees tested.
\citet{jfrDetectionAndDTM} had five test sites, containing 113 trees.
In \citep{tsubouchi2014forest}, testing was done in one forest with no branches or vegetation-occluding trunks, validating against nine measured trees.
\citet{calderscomp} had the most complete dataset, with 10 test sites consisting of a circle of \SI{15}{\m} radius containing a total of 331 trees.
While \citet{tang2015slam} did not assess their \ac{DBH} measurements, they measured the position of 224 trees with a total station along one trajectory.
While the results in \citep{seiki2017backpack} were encouraging, the validation was conducted on one forest site with seven reference trees.
\citep{marek} evaluated their work in one site under near-perfect conditions; no branches occluding the stem and no ground vegetation causing occlusion.

\section{Methods}
\label{subsec:methods}

We describe here our data processing pipeline, starting with map generation, tree segmentation and  determining breast height. Finally, we present our different diameter estimation algorithms.

\subsection{Iterative closest point mapping}
\label{sec:mapping}

Our 3D mapping method relies on a modified version of \texttt{ethz-icp-mapping} \citep{icpmapping}, which uses the \ac{ICP} algorithm as the registration solution.
\ac{ICP} takes as input a reading point cloud $\mathbf{Q} \in \mathbb{R}^{3 \times m}$ (i.e., the current lidar view of the robot) containing $m$ points, a map point cloud $\mathbf{M}' \in \mathbb{R}^{3 \times l}$ containing $l$ points, and an initial pose estimate $\mathbf{\hat{T}} \in \text{SE}(3)$ to estimate the pose of the robot in the map.
To compute the initial estimate, we used an \ac{EKF} to fuse our \ac{IMU} and wheel odometry.
Reading point clouds are filtered for dynamic elements and maps are uniformly downsampled to keep computation time reasonable.
The mapping was not performed in real time, as we prioritized map quality over computation time.
The algorithm is described in \autoref{alg:mapper}.

\begin{algorithm}
 \caption{The ICP mapping algorithm used in this paper.}
 \label{alg:mapper}

\begin{algorithmic}[1]
\Procedure{Mapping}{point clouds $\mathbf{Q}_{0:t}$, odometry $\mathbf{O}_{0:t}$}
\State $\mathbf{T}_0 \gets \mathbf{1}$ \Comment{Initial pose is at the origin, no rotation}
\State $\mathbf{M}'_0 \gets \mathrm{inputFilters}(\mathbf{Q}_0)$ \Comment{Initial map, compute normals}
\For{i = 1..t}
\State $\mathbf{\hat{T}}_{i} = \mathbf{O}_{i-1}^{-1} \mathbf{O}_{i} \mathbf{T}_{i-1} $
\Comment{The initial estimate is the last pose combined with odometry}
 \State $\mathbf{Q}'_i = \mathrm{inputFilters}(\mathbf{Q}_i)$ \Comment{Compute normals}
 \State $\mathbf{T}_i = \mathrm{icp}(\mathbf{Q}'_i, \mathbf{M}_{i-1}, \mathbf{\hat{T}}_i)$ \Comment{Refine pose estimate with \ac{ICP}}
 \State $\mathbf{M}_i = \begin{pmatrix} \mathbf{M}_{i-1} & | & \mathbf{T}_i \mathbf{Q}'_i \end{pmatrix}$ \Comment{Add points to map}
  \State $\mathbf{M}'_i = \mathrm{reduceDensity}(\mathbf{M}_i)$ \Comment{Aim for a density of one point per \SI{8}{\cm^3}}
  \EndFor
  \State \textbf{output} final map $\mathbf{M}'_t$ and trajectory $\mathbf{T}_{0:t}$
\EndProcedure
\end{algorithmic}
\end{algorithm}

\subsection{Point selection for \ac{DBH} estimation}
\label{sec:slice-description}
The first step in estimating \ac{DBH} from a 3D map is to segment trees.
Although automatic methods exist~\citep{jfrDetectionAndDTM, computree}, we chose to perform the tree segmentation manually.
Our motivation is to validate diameter estimation methods also on less visible trees regardless of segmentation quality.
Our manual segmentation comes in the form of 3D bounding-boxes around trunks, which were manually adjusted.
These bounding-boxes can include branches and noise, which will be outliers stressing the \ac{DBH} estimation methods.

To estimate the \ac{DBH}s, we had to locate the breast height of each tree, defined as \SI{1.3}{\m} above ground level.
This implies estimating the ground level at each tree location in the point cloud using a \ac{DTM}.
Several algorithms have been designed for this purpose~\citep{jfrDetectionAndDTM}, from which we chose the raster-based method.
The ground height for a given tree was the value of the \ac{DTM} given the $(x,y)$ position of the center of the manually drawn bounding box.
Then, we selected every point in the tree bounding box which was between $h/2$ below breast height and $h/2$ above, where $h$ represents a section thickness.
Selecting this thickness $h$ was a trade-off between inducing an error from the change in diameter along a tree's height and the fact that cylinder fitting performs better as more points are available.
Points resulting from this selection are colored in \autoref{fig-OverviewMap}.
Those points were finally used to estimate the \ac{DBH} by one of the cylinder fitting methods described below.

\subsection{Least square cylinder fitting}
\label{sec:cylinder-fitting}
As commonly done \citep{jfrDetectionAndDTM, seiki2017backpack, computree}, we formulate tree diameter estimation as cylinder-fitting.
Fitting cylinders to point clouds is a fairly well studied problem \citep{lukacs1998faithful}.
Let $\mathbf{P} = \begin{pmatrix} \mathbf{p}_1 & \mathbf{p}_2 & \dots & \mathbf{p}_n \end{pmatrix} \in \mathbb{R}^{3 \times n}$ be the slice in our point cloud described in \autoref{sec:slice-description} and containing $n$ points belonging to one tree.
We also have $\mathbf{N} = \begin{pmatrix} \mathbf{n}_1 & \mathbf{n}_2 & \dots & \mathbf{n}_n \end{pmatrix} \in \mathbb{R}^{3 \times n}$, which are the surface normals for each $\mathbf{p}_i$.
We used the spectral decomposition of the covariance matrix from the $q$-nearest neighbors for each $\mathbf{p}_i$ to estimate $\mathbf{N}$.
The eigenvector associated with the smallest eigenvalue of this matrix is the direction of least variance, corresponding to the estimated normal of the surface.
A cylinder used to fit $\mathbf{P}$ can be represented in multiple ways.
For this work, we parametrize a cylinder as $(\mathbf{a},\mathbf{c},r)$ where $\mathbf{a} \in \mathbb{R}^3 : \left\Vert \mathbf{a} \right\Vert_2 = 1$ is the cylinder axis direction, $\mathbf{c} \in \mathbb{R}^3$ is any point on the cylinder axis and $r \in \mathbb{R}^+$ is the cylinder radius.
This parametrization has seven parameters, with one degree of freedom removed from the axis norm constraint; the last degree of freedom can be removed by imposing $\mathbf{a} \cdot \mathbf{c} = 0$ which comes naturally when solving for $\mathbf{c}$ in the next cylinder fitting method presented.
From there, we investigated four methods to find those parameters from the point cloud $\mathbf{P}$.

\textbf{1) Finding the axis using surface normals} ---
The \emph{linear least square} method ($A_{LLS}$)~\citep{li2018supervised} needs the surface normals $\mathbf{N}$.
This axis-finding method is based on the fact that if $\mathbf{P}$ and $\mathbf{N}$ represent a perfect cylinder, then all normals $\mathbf{n}_i$, $i = 1 \dots n$ will lie on a plane passing through the origin for which the normal will be $\mathbf{a}$.
Therefore, finding the optimal axis $\mathbf{a}^*$ for a cylinder can be done by solving
\vspace{-7pt}
\begin{equation}
    \mathbf{a}^* = \argmin_{\mathbf{a}} \left\Vert \mathbf{N^{\intercal}a} \right\Vert_2 .
\vspace{-7pt}
\end{equation}
As it turns out, $\mathbf{a}^*$ is the third right singular vector of the singular value decomposition of the matrix $\mathbf{N}$. A useful property of the $A_{LLS}$ method is that it is linear.
We can then project $\mathbf{P}$ on a plane perpendicular to $\mathbf{a}^*$; the resulting 2D point cloud can be used to fit a circle using any known method discussed in the next paragraph.
This circle fitting method will find the remaining parameters  $r$ and $\mathbf{c}$.
Another approach ($A_N$) in finding the cylinder axis is assuming that the tree is perfectly vertical, leading to the simplification $\mathbf{a} = \begin{pmatrix} 0 & 0 & 1 \end{pmatrix}$. This approach was employed in~\citep{tsubouchi2014forest, marek}.

\textbf{2) Circle fitting algorithms} ---
Once the axis $\mathbf{a}$ of a cylinder is known, one can project the points in $\mathbf{P}$ on a plane perpendicular to this axis and then fit a circle to find the radius $r$ and center $\mathbf{c}$.
This can be done using \emph{iterative} or \emph{algebraic} methods.
The iterative methods consist of minimizing the sum of squares of the point-to-circle distance using iterative methods.
Consequently, they are prone to local minima issues.
The algebraic methods do not rely on iterative methods, but rather analytically solve the problem of circle fitting using an approximation of point-to-circle distance~\citep{pratt1987direct}.
In our experiments, we used an algebraic fit called \emph{Hyper}, abbreviated as $H$, introduced in \citep{al2009error}.
In their paper, the authors prove that they have a non-biased fit, as opposed to~\citet{pratt1987direct}, in the case of incomplete circle arcs.

\textbf{3) Non-linear least square cylinder fitting} ---
This method, presented by \citet{lukacs1998faithful}, uses non-linear optimization to estimate the complete cylinder parameters.
It relies on a point-to-cylinder distance:
$d(\mathbf{p}_i; \mathbf{a}, \mathbf{c}, r) = \left\Vert \left(\mathbf{p}_i - \mathbf{c} \right) \times \mathbf{a} \right\Vert_2 - r.$
To find the cylinder, one then solves
\vspace{-7pt}
\begin{equation}\label{eq-least-square-cylinder}
 \mathbf{a}^*, \mathbf{c}^*, r^* = \argmin_{\mathbf{a}, \mathbf{c}, r} \hspace{3pt} \sum_{i = 1 \dots n} d^2(\mathbf{p}_i; \mathbf{a}, \mathbf{c}, r) \underbrace{+ \sum_{i = 1 \dots n}(\mathbf{n}_i \cdot \mathbf{a})^2}_{\text{Normals loss (optional)}}.
\vspace{-7pt}
\end{equation}
The second sum is optional, but can be added to the minimization to penalize cylinders which do not fit $\mathbf{N}$ well.
To the best of our knowledge, this penalty has not been described elsewhere in the literature.
For this paper, the original method without normals will be called $C_{NLS}$, while the minimization with the extra penalty will be called $C_{NLSN}$.
We can solve this optimization problem in an unconstrained manner, by converting the problem to the cylinder parametrization from \citet{lukacs1998faithful}.

\textbf{4) Multiple cylinders voting} ---
We can fit multiple cylinders to the tree slice to improve the estimate robustness.
In this case, we divide our tree slice vertically to form $n_{cyls}$ point clouds, and fit a cylinder to each one.
Then, one can choose the median ($V_{median}$) or the mean ($V_{mean}$) of the diameter of the cylinders as the \ac{DBH}.

\section{Experimental Setup}
\label{sec:exp_setup}

For each tree in our test sites, a forest technician identified all species and measured the diameter of trees using a specialized diameter tape.
This information was engraved on a small metal marker attached to each tree.
The only criteria for tree inclusion in the dataset was that
\begin{enumerate*}
	\item its diameter was greater than \SI{2}{\cm} and
	\item the tree was standing.
\end{enumerate*}
We generated initial 3D maps from our robot observations, and then segmented every individual tree in these maps, assigning an ID to each tree.
Afterwards, a stem map (i.e., a two-dimensional plot of the position of every individual tree and its ID) was generated and printed on paper.
We then used this stem map in the field to associate each tree to its ID, and then recover the measurement made by the technician.
Unfortunately, even differential GPS cannot be used to localize trees sufficiently precisely for this task, due to canopy interference.
In the end, the information for each tree included
\begin{enumerate*}
    \item an individual ID,
    \item its position in the 3D map in the form of a bounding box,
    \item its ground-truth \ac{DBH}, and
    \item its species.
\end{enumerate*}

%
We used a \emph{Clearpath Husky A200} mobile robot to map the different forest sites.
Its skid drive makes it appropriate for navigating rough forested environments and emulating forest machinery.
The robot was equipped with a \emph{Velodyne HDL32} lidar, an \emph{Xsens MTI-30} \ac{IMU} and wheel encoders for odometry.
All processing was performed offline on a workstation with an AMD Ryzen 1700 and 64 GB of RAM.

\begin{figure}[ht!]
    \centering
    \begin{subfigure}[b]{0.49\textwidth}
        \centering
        \begin{overpic}[width=\linewidth, trim={0 17cm 0 0}, clip]{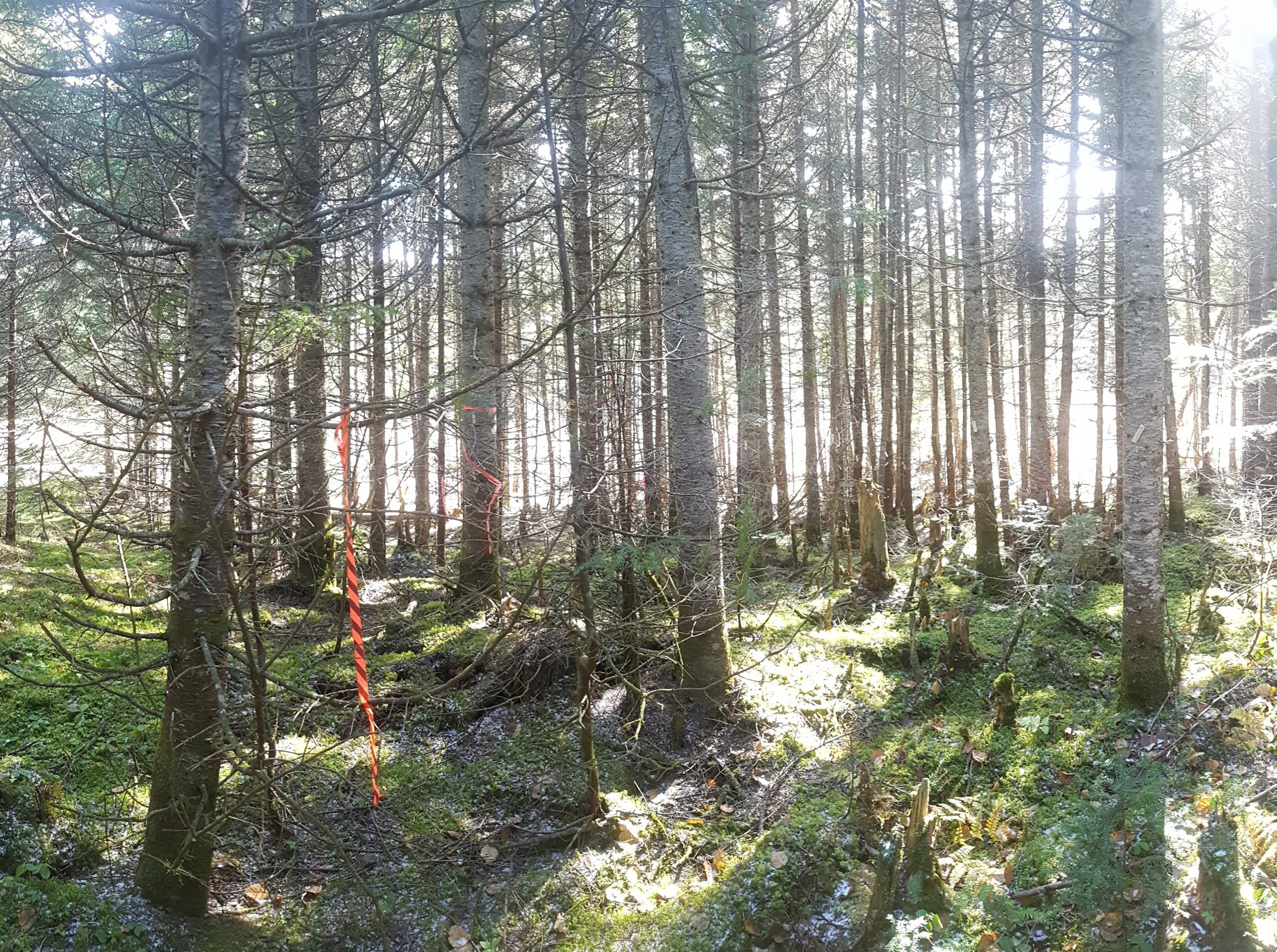}
          \put (5.5, 43.5) {\Large \color{black} \textbf{\textsc{Young}}}
          \put (5, 44) {\Large \color{white} \textbf{\textsc{Young}}}
        \end{overpic}
    \end{subfigure}
    ~
    \begin{subfigure}[b]{0.49\textwidth}
        \centering
        \begin{overpic}[width=\linewidth, trim={0 26cm 0 0}, clip]{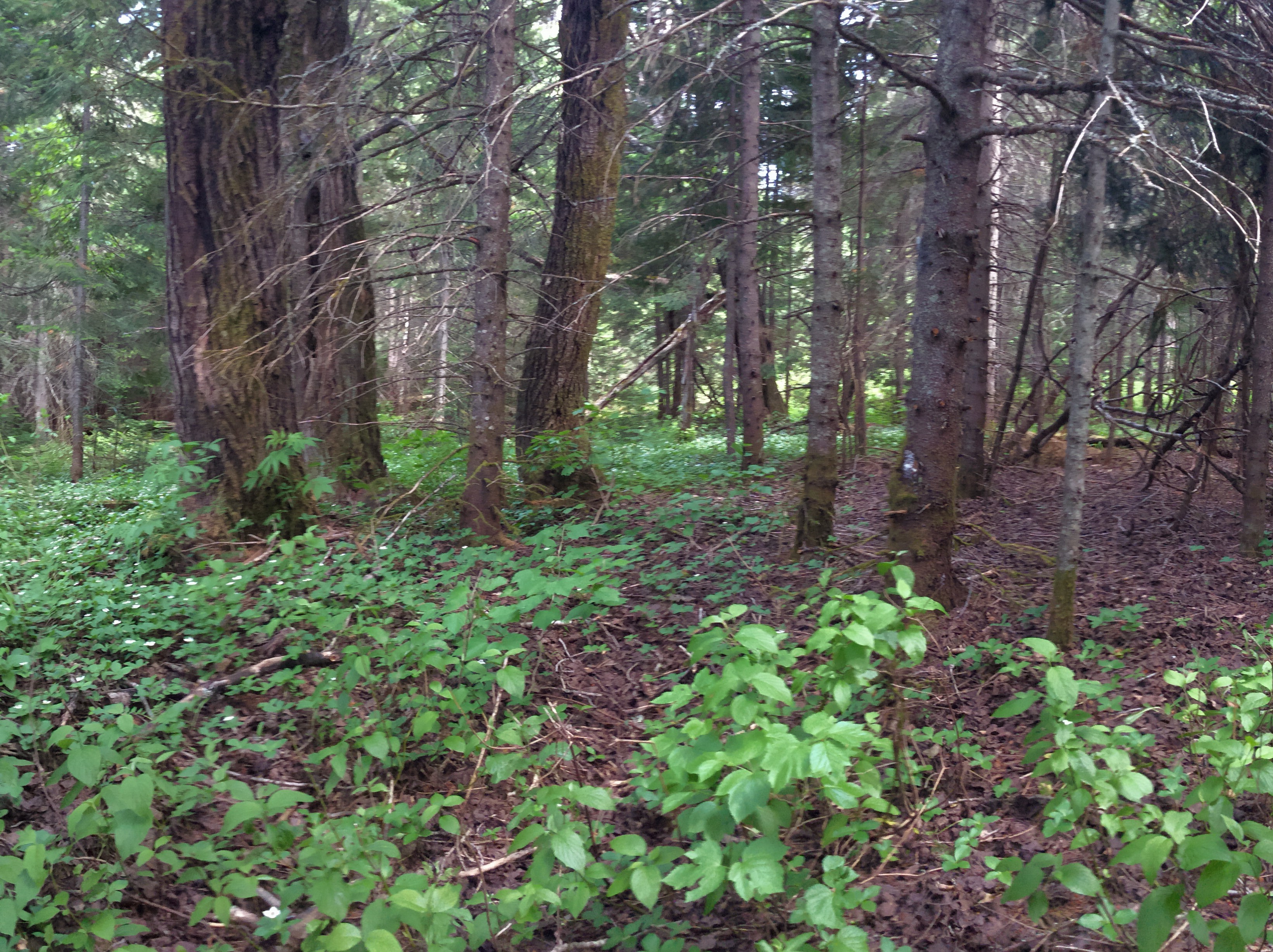}
          \put (5.5, 43.5) {\Large \color{black} \textbf{\textsc{Mixed}}}
          \put (5, 44) {\Large \color{white} \textbf{\textsc{Mixed}}}
        \end{overpic}
    \end{subfigure}
    \\ \vspace{5pt}
    \begin{subfigure}[b]{0.49\textwidth}
        \centering
        \begin{overpic}[width=\linewidth, trim={0 10cm 0 1cm}, clip]{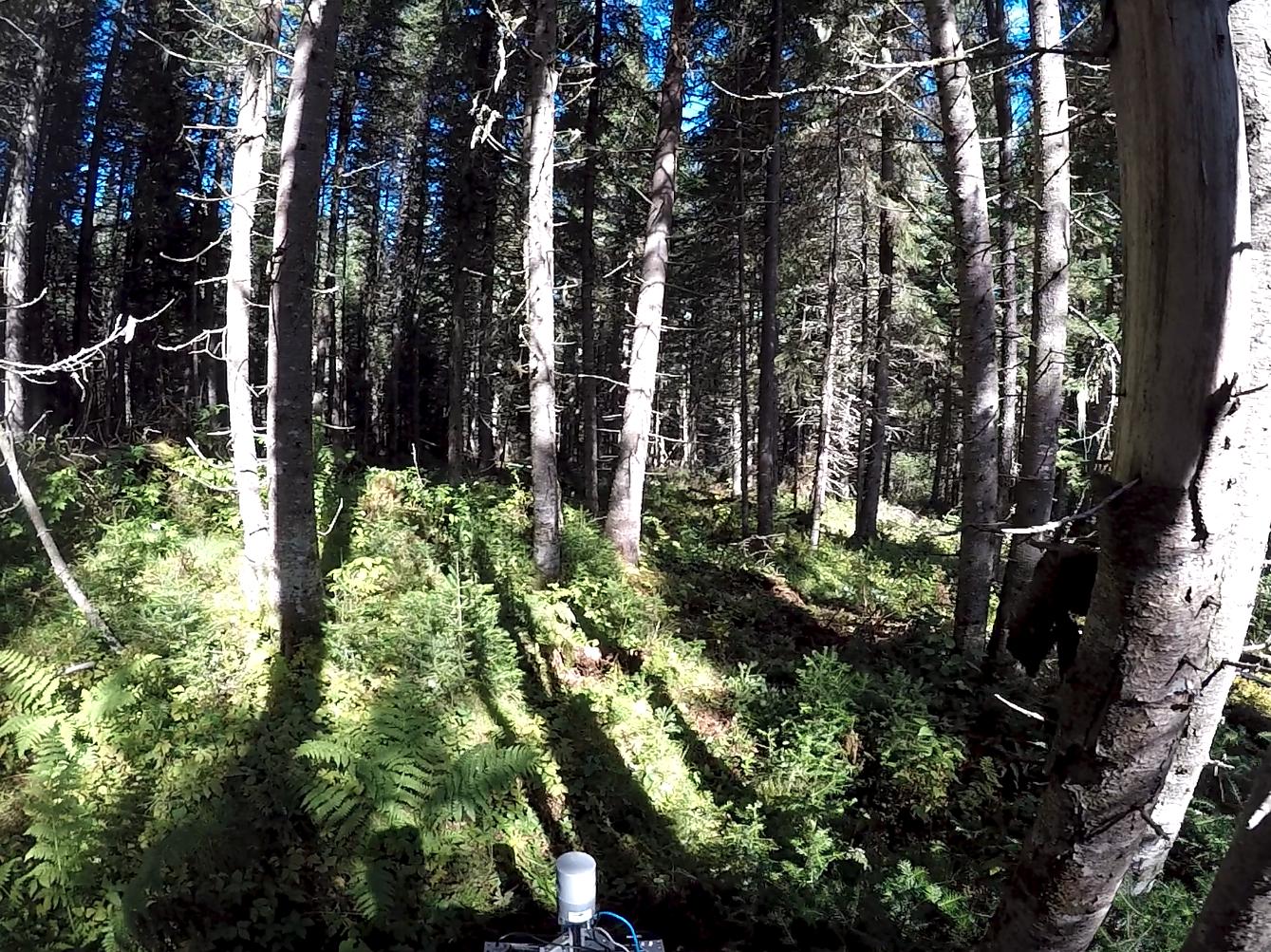}
          \put (5.5, 43.5) {\Large \color{black} \textbf{\textsc{Mature}}}
          \put (5, 44) {\Large \color{white} \textbf{\textsc{Mature}}}
        \end{overpic}
    \end{subfigure}
    ~
    \begin{subfigure}[b]{0.49\textwidth}
        \centering
        \begin{overpic}[width=\linewidth, trim={0 6cm 0 5cm}, clip]{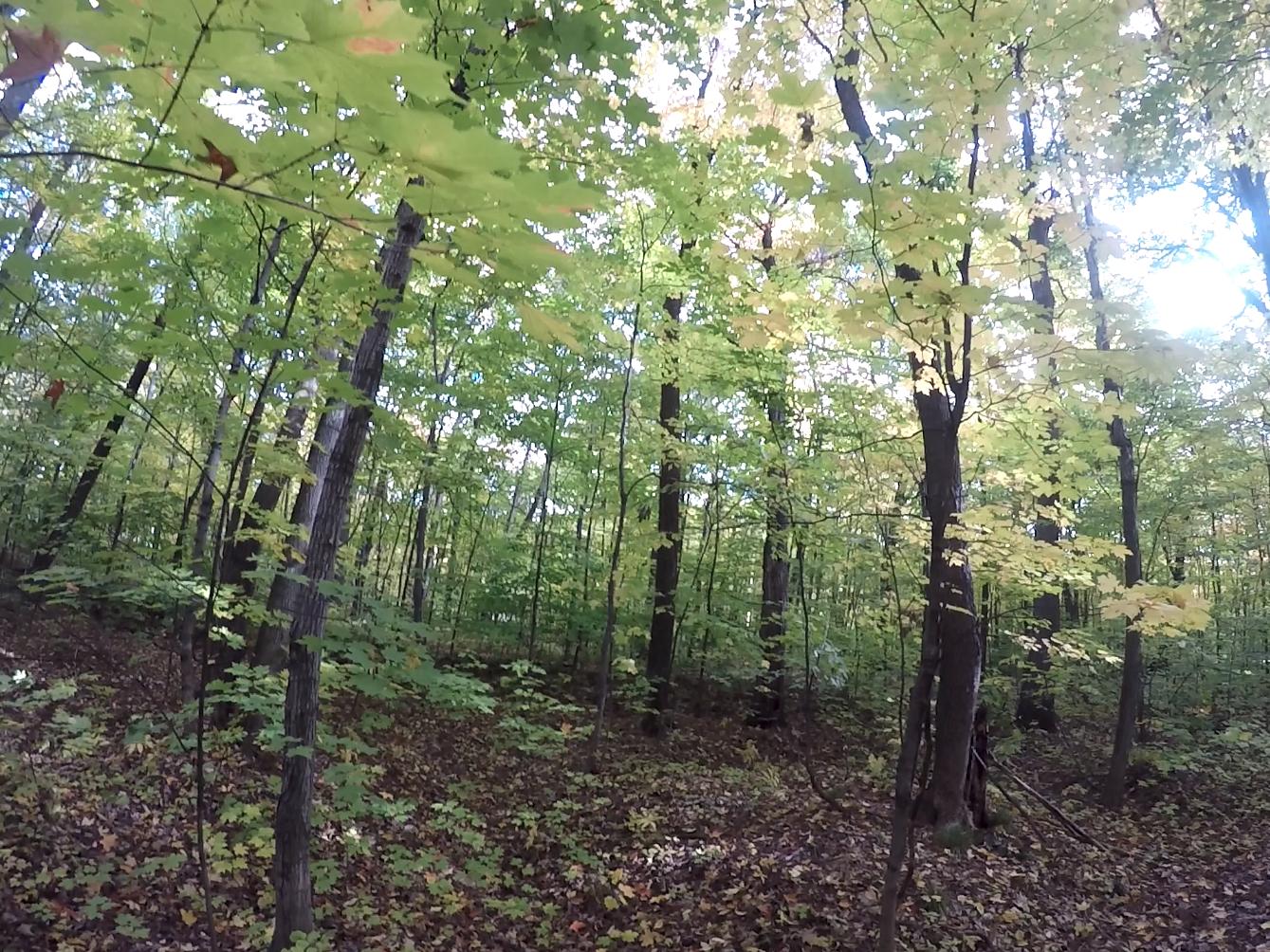}
          \put (5.5, 43.5) {\Large \color{black} \textbf{\textsc{Maple}}}
          \put (5, 44) {\Large \color{white} \textbf{\textsc{Maple}}}
        \end{overpic}
    \end{subfigure}
    \caption{Photos of our test sites. One can see the variety in composition and density; from tightly planted balsam firs in \textsc{Young} to well spaced mature spruce in \textsc{Mature}. Also visible is the rough ground of \textsc{Young} and \textsc{Mature}, compared to that of \textsc{Mixed} and \textsc{Maple}.}
    \label{fig-site-pictures}
\vspace{-17pt}
\end{figure}
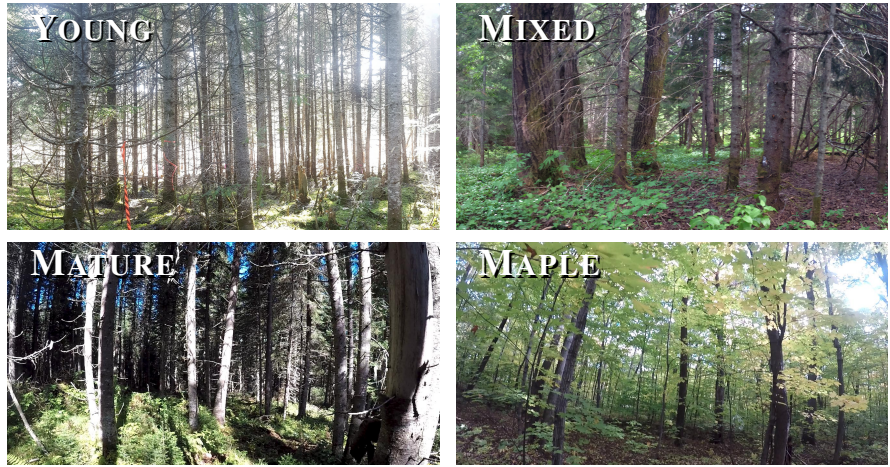

\subsection{Experimental sites}
\begin{figure}[ht!]
    \centering
    \includegraphics[width=\linewidth]{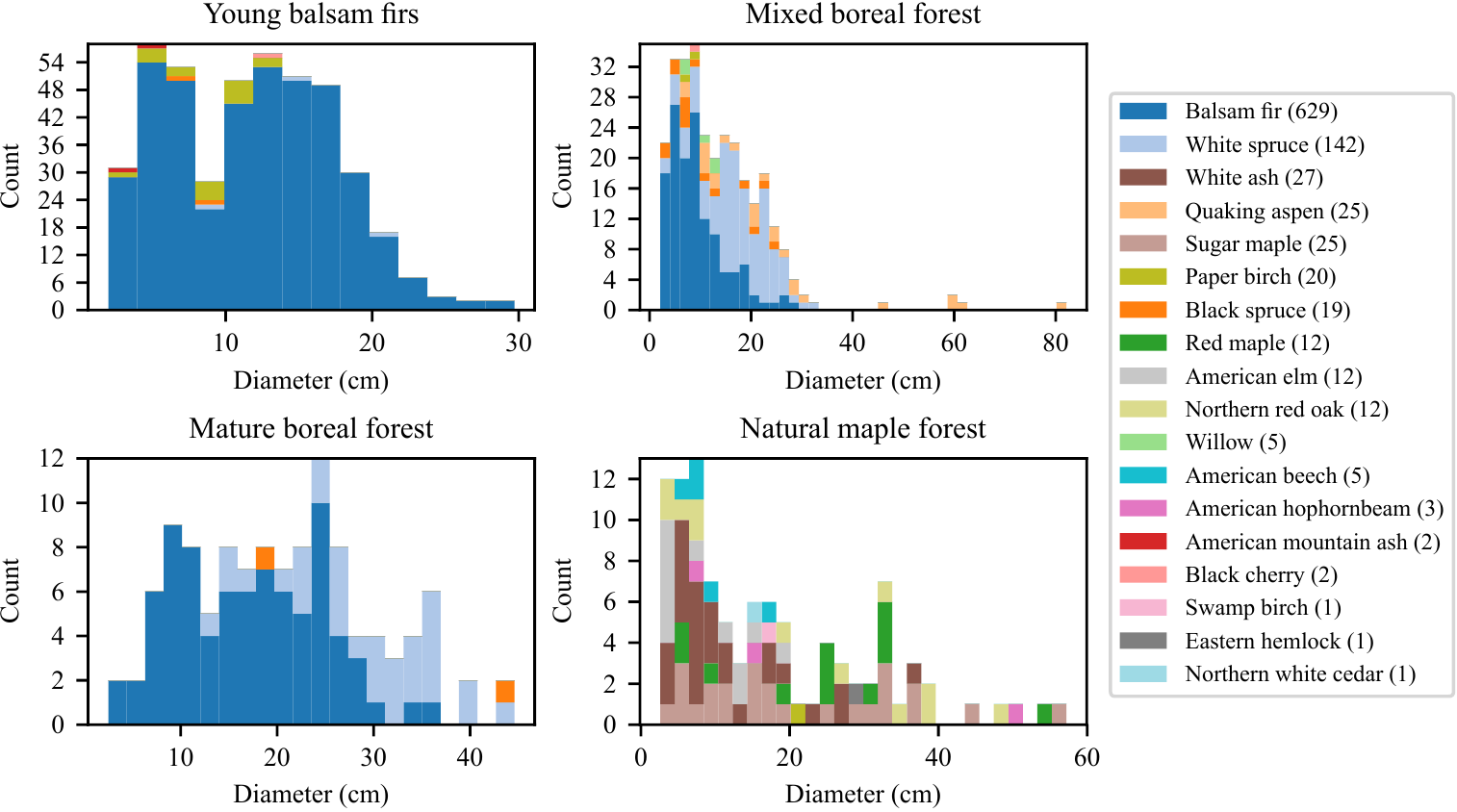}
    \caption{Diameter and species distribution for our test sites, with the tree count by species in parenthesis. Notice the species and diameter diversity in the different sites. This allowed us to verify the impact of species on diameter estimation, as bark texture impacts the point cloud produced by the lidar. Trees of less than 10 cm were segmented and measured, but were not used in this paper.}

    \label{fig-site-composition}
\vspace{-5pt}
\end{figure}

We collected data on four significantly different sites, scanning in total 1.4 hectares. We manually measured and marked 943.
From these, 588 were above \SI{10}{\cm} in \ac{DBH} (trees below this threshold are not considered of commercial value) and kept for our study.
We chose sites that were different in terms of age, composition (see \autoref{fig-site-composition}), density and topography (see \autoref{fig-site-pictures}), to identify how these factors could affect our diameter estimation and robot mapping.
The first three sites were located at For\^et Montmorency, owned by Laval University.
The last one was located on the University campus, as For\^et Montmorency contains little deciduous forest.
We tested a number of trajectories for each site, to see their impact on diameter estimation.
The different trajectories were placed in a common coordinate system for each site, using the approach proposed in~\citep{tlr}.
For our analysis, each tree observed in a given trajectory is considered as an individual tree observation.
Therefore, we have in our dataset 2 to 4 observations for each tree.
By far, these four sites represent the largest \ac{DBH} dataset from mobile lidar in the literature. We describe them below.

\textbf{1) Young balsam firs} (\textsc{Young}) ---
Despite being a plantation, the topography of this site was very rough, with a \SI{30}{\%} incline and mossy soil. The robot experienced frequent slippage, thus affecting odometry.
There were a lot of lower branches occluding the trunks at breast height, which could affect our measurements.
The trees were tightly planted, resulting in reduced visibility for the lidar.
These factors make this site challenging both for perception and navigation.
This site was mostly composed of balsam firs with some paper birch
and measured \SI{30}{\m} $\times$ \SI{35}{\m}.
Two trajectories were performed on this site: one big loop around the site, and a longer one where we made a loop around the site but also crossed the site in the middle.

\textbf{2) Mixed boreal forest} (\textsc{Mixed})
Despite being generally flat, this site had a lot of branches on the ground, making navigation difficult.
The understory vegetation was also very dense in some places, thus limiting visibility.
It was diverse in terms of tree species and age, consisting mainly of quaking aspens, balsam firs and spruce.
The site was \SI{50}{\m} $\times$ \SI{30}{\m}.
We conducted three trajectories: the first one was a simple loop around the site, while the other two tried to be more exhaustive.

\textbf{3) Mature boreal forest} (\textsc{mature})
There was a \SI{10}{\%} incline, not as sharp as \textsc{Young}, and very irregular ground.
This site had big trees with non-occluded trunks.
Being a mature forest, there was a fair number of fallen trees which could block the robot.
The site was mostly composed of balsam fir as well as white and black spruce.
This site measures \SI{30}{\m} $\times$ \SI{40}{\m}.
Two trajectories were performed, similarly as in \textsc{Young}.

\textbf{4) Mature natural maple forest} (\textsc{Maple})
It was flat and easy to navigate, with very few obstacle on the ground.
Consequently, we drove the robot at a faster speed (1 m/s) for much of the trajectories.
It was a mature decideous natural forest, composed mainly of sugar and red maple.
The site was \SI{100}{\m} $\times$ \SI{100}{\m} and contained upwards of 1000 trees.
To reduce the ground-truth labor, we randomly selected 100 trees which we would measure.
Two trajectories, similar to \textsc{Young} and \textsc{Mature}, were performed at the beginning of October with the leaves on the trees.
The same trajectories were repeated at the beginning of November with no leaves left, to study the potential impact of leaves on diameter estimation.

\section{Results and discussion}

\subsection{Comparing diameter estimation approaches}
We tested different combinations of the methods presented in \autoref{sec:cylinder-fitting}, to determine the best one.
First, we tested the combination of $A_N$ and $H$, which can be summarized as fitting a circle to the $xy$-coordinates of the tree slice.
Similarly, we tested $A_{LLS}$ and $H$.
Also, we tested combining both methods above with $C_{NLS}$ and $C_{NLSN}$, using both of the first as initial estimates for the two later.
All six resulting methods were combined with either $V_{median}$ and $V_{mean}$, resulting in 12 sets of results.
All combinations used RANSAC as the outlier rejection method, with tolerance $\varepsilon$.
We limited this comparison for trees observed at a distance closer than \SI{10}{\m}.
As will be shown in \autoref{sec:impact-min-distance}, this minimal observation distance has too large of an impact on the estimation of \ac{DBH}, and we consider accurate \ac{DBH} measurement from this distance currently unfeasible.
This meant discarding 143 out of our 1458 tree observations.
We tested all of the following hyperparameter values: $q \in \{15, 20, 25\}$, $n_{cyls} \in \{1, 2, 3, 4, 5\}$, $h \in \{20,30,40,50,60\}$ cm, and $\varepsilon \in \{1,2,3\}$ cm.

Because $V_{mean}$ consistently underperformed $V_{median}$ in all of our tests, its performance is not reported in \autoref{tab:DBH_Results}.
The inferior performance of $V_{mean}$ was also confirmed by the fact that the number of vertical slices $n_{cyls} = 1$ was always the best choice in our hyperparameters exploration with $V_{mean}$, meaning that not using $V_{mean}$ was preferable to using it.

\newcolumntype{R}{>{\raggedleft\arraybackslash}X}
\begin{table}
  \centering
  \caption{Results of our diameter estimation methods combined with $V_{median}$ on our dataset (11 trajectories on four sites).
    The fail rate is the proportion of trees where there were not enough points, or the error was above \SI{20}{\cm}. %
    We treated those failures as outliers; they were not used to compute the RMSE or bias.
  The result shown for each method was computed using the best combination (in terms of RMSE)
  of hyperparameters.
  Because \emph{Hyper} is used for all of the methods, it is removed from the first row.
  }
  \begin{tabu}{ X R R R R R R }
    \toprule
    & $A_{LLS}$ & $A_{N}$ & $A_{LLS} + C_{NLS}$ & $A_{N} + C_{NLS}$ & $A_{LLS}\hspace{-1pt}+\hspace{-1pt}C_{NLSN}$ & $A_{N} + C_{NLSN}$ \\ \midrule
    RMSE (cm)          &  5.08 & 4.41        &  3.76 &  \textbf{3.45}        &  3.86 & 3.66 \\
    Bias (cm)          & -0.95 & 0.72        & -0.62 & -0.41        & -0.18 & \textbf{0.00} \\
    Fail rate (\%)     & 11.41 & 13.61       &  \textbf{6.23} &  6.46        &  6.54 & 7.53 \\ \hline
    $q$                & 25    & \textit{N/A} &  25   &  \textit{N/A} &  25   & 20 \\
    $n_{cyls}$          & 1     & 3           &  5    &  5           &  3    & 5 \\
    $h$ (cm)           & 20    & 60          &  60   &  60          &  40   & 50 \\
    $\varepsilon$ (cm) & 3     & 3           &  2    &  2           &  2    & 1 \\
    \bottomrule
  \end{tabu}
  {\scriptsize\textit{Legend}: $A_{LLS}$--lin. l.-s. axis, $A_{N}$--vertical axis, $C_{NLSN}$--non-lin. l.-s. with normals, $C_{NLSN}$--without normals.}
  \label{tab:DBH_Results}
\vspace{-15pt}
\end{table}

One can see that the best performing method is $A_N + C_{NLS}$ and that the worst is $A_{LLS}$.
One conclusion from our comparison is that $V_{median}$ leads to better results.
All of the methods, except $A_{LLS}$, performed better when $n_{cyls}$ was larger than one.
Surprisingly, using a pure vertical tree axis ($A_N$) performed better than trying to take into account the stem direction ($A_{LLS}$), even as an initial estimate to non-linear cylinder fitting.
This suggests that $A_{LLS}$ is not precise enough to estimate the stem direction accurately in noisy mobile lidar point clouds.
However, the vast majority of our trees grew vertically in our dataset, thus there may be a bias favoring $A_N$.

\begin{SCfigure}
    \centering
    \includegraphics[width=0.6\textwidth]{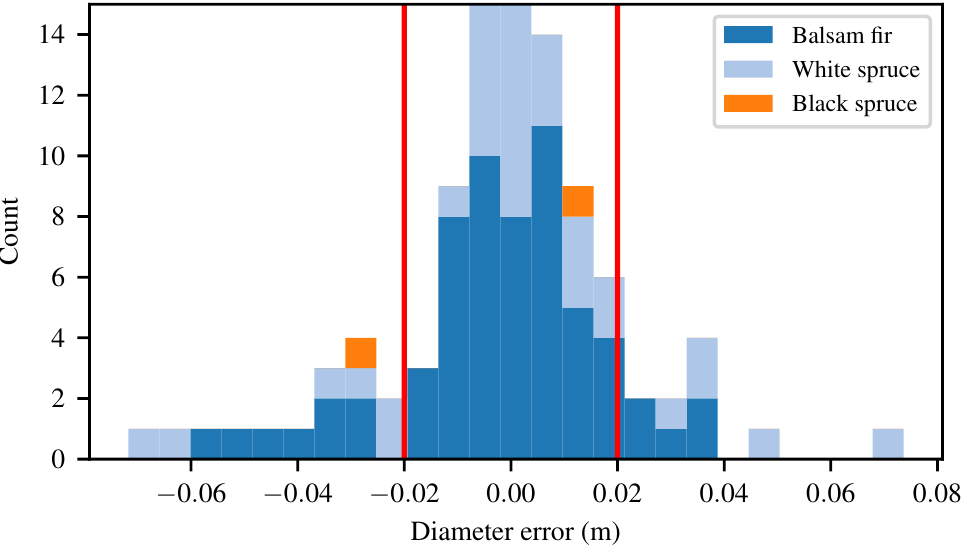}
    \caption{
      Error distribution over 94 trees for our best performing site, \textsc{Mature}, using the second trajectory.
      In red is the \SI{2}{\cm} error limit.
      Overall, the \ac{RMSE} is \SI{2.16}{\cm}, the bias is \SI{0.15}{\cm}, and the \ac{RMSE} of trees observed closer than \SI{8}{\m} (all but 4 trees, which are at the tail ends of the distribution is this figure) is \SI{2.04}{\cm}. %
      The hyperparameters for these results were $\varepsilon=$\SI{1}{\cm}, $h=$\SI{40}{\cm}, $n_{cyls}=3$ and $q=15$.}
    \label{fig:best_results}
\vspace{-15pt}
\end{SCfigure}

Our best performing site was \textsc{Mature}, with its well-spaced trees and visible trunks.
\autoref{fig:best_results} gives an example of the error distribution achieved in one trajectory with $A_{LLS} + H + C_{NLSN} + V_{median}$ in these ideal circumstances. %

\subsection{Factors impacting \ac{DBH} estimation}
\label{sec:impact-min-distance}
\begin{SCfigure}
    \centering
    \caption{Error vs. minimal observation distance for \textsc{Maple}.
      We can see that the error spread (in terms of interquartile ranges) is higher in leaf-on conditions, from 0 to 10 m of observation distance. After \SI{10}{\m}, the error becomes too great for forest inventory purposes. Note the degradation after \SI{20}{\m} in leaf-on conditions, due to the lack of points.}
    \includegraphics[width=0.6\textwidth]{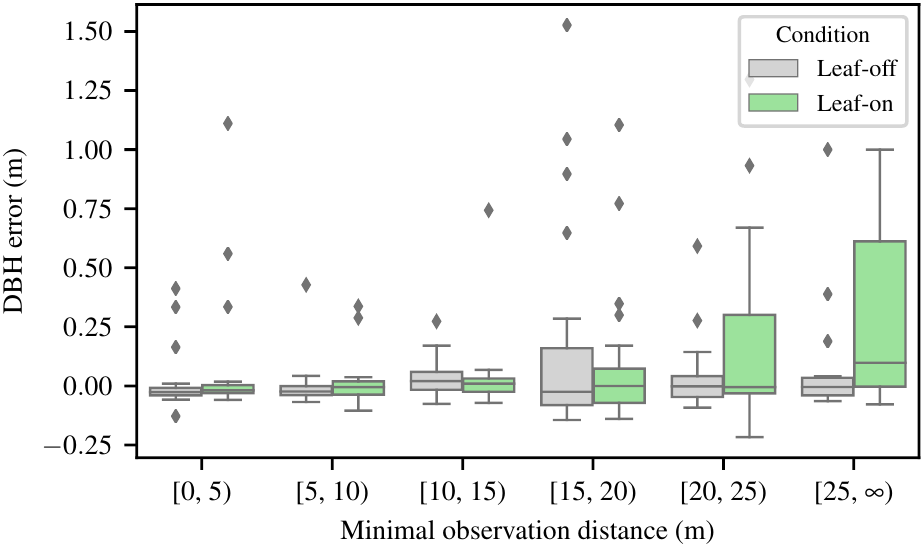}
    \label{fig:distance}
\vspace{-18pt}
\end{SCfigure}

The following statistics were generated using the $A_{N} + H + C_{NLS} + V_{median}$ method, but similar observations can be made for others.
We tried to identify possible factors impacting the estimation of \ac{DBH}.
For instance, we can see in \autoref{fig:distance} the impact of minimal observation distances and presence of foliage on the error distribution, for the \textsc{Maple} dataset.
This minimal observation distance represents how close the robot was driven from a tree.
We observe that the error becomes too high (i.e., more than \SI{10}{\cm} of \ac{RMSE}) for trees to which the robot has not gotten closer than \SI{10}{\m}, particularly when trees have foliage; again, this is similar for the other three sites.
The effect of foliage could be due to increased localization error incurred during the map creation process of \autoref{sec:mapping} or reduced trunk visibility.

We also observed that the localization error can result in a reduced diameter estimation accuracy.
In \textsc{Mature}, the trajectory that looped around the site resulted in significant localization error, as stem sections of the same trees observed at the beginning of the trajectory were misaligned with observations made at the end of the trajectory.
This caused a bias of \SI{-1.01}{\cm} for the loop, while the other two trajectories had a respective bias of \SI{-0.31}{\cm} and \SI{0.03}{\cm}.
This trajectory was the only one in our dataset with such visible localization error.

Finally, another factor impacting the estimation of \ac{DBH} is species, which we attributed to the influence of bark roughness.
For example, the big red maples in \textsc{Maple} had a bias of \SI{-5}{\cm}, the quaking aspens of \textsc{Mixed} had a bias of \SI{-1.61}{\cm} while the very smooth balsam firs overall had limited bias (\SI{0.47}{\cm}), after removing the results from the problematic trajectory mentioned above.
The same effect of bark texture on \ac{DBH} measurements has also been observed by \citet{calderscomp}.

\subsection{Lessons learned}
During the course of this work, we gained significant experience in field deployment of mobile robots in forests.
Here are some lessons and recommendations that could be useful to anyone interested in deploying robots in forests for 3D mapping and inventory purposes, as well as more sophisticated experiments where observing trees is important.

\begin{itemize}
\item Ground roughness, such as branches, irregular ground or other obstacles, did not seem to have an significant impact on our mapping and diameter estimation, as our two best performing sites, \textsc{Mature} and \textsc{Young}, were also the ones where the robot had the most trouble navigating.
\item Getting close to each tree is essential for \ac{DBH} estimation: our experiments shows that diameter estimation performance degrades rapidly for observations beyond \SI{6}{\m} and becomes unusable after \SI{10}{\m}.
  This could be due to propagation of the orientation estimation error during the map building process, lidar beam width or point density reduction.
  In each site, more exhaustive trajectories always performed better than simple loops.
  For example in \textsc{Mixed}, we even observed a negative bias caused by localization error in the simple loop.
\item Our results in \textsc{Maple} were affected by the robot speed. At 1 m/s, the platform moves \SI{10}{\cm} during each lidar scan, which is ignored by ICP. This limitation of the algorithm led to poorer results in what, we thought, should have been the best performing site.
  Solving this issue by inferring the movement of the platform during one scan, such as done by \citet{loam}, is important if we want this approach to work in fast moving forest robots.
  It would also allow for faster data acquisition, as well as possibly lead to more accurate \ac{DBH} estimation overall.
\item Mobility in forests is challenging; we pushed our robot to its limit.
  Using continuous tracks would be ideal, as it is used for most forest machinery. In \textsc{Mature} we had to slightly alter the environment for our robot by cutting three fallen trees (which are abundant in mature forests), while other sites were navigable with no modifications.
  From a pure forest inventory standpoint, a backpack-mounted or handheld sensor system would be better suited, but it remains interesting nonetheless to study robot perception in such difficult conditions.
\end{itemize}

\section{Conclusion}
\label{sec:Conclusion}
In this paper, we presented an ICP mapping approach in forests, and demonstrated that it produced maps that were accurate enough to perform tree diameter estimation, especially in mature, well-spaced forests where we reached an accuracy of \SI{2.04}{\cm}.
We also identified key challenges to address in robot mapping in forests: dealing with rough tree bark, reduced visibility in dense forest and estimating platform motion during one scan.
We compared multiple diameter estimation methods, and concluded that fitting multiple cylinders using \emph{Hyper} circle fitting combined with non-linear cylinder fitting, and taking the median of the diameters of those cylinders works best.
All of our methods were validated in the most extensive dataset of \ac{DBH} measurement from mobile lidar in the literature.

\subsection{Future work}
3D mapping opens the door for future automation of forestry equipment.
The next step would be localizing a forest harvester in our 3D point cloud in real time.
More work is needed in this direction as we did not attempt getting our mapping algorithm to run in real time while producing maps that were accurate enough to measure the \ac{DBH} of trees.
Trying to use trees as landmarks to take some pressure off \ac{ICP} could be a solution.
Furthermore, integrating work by \citet{mathieu} to perform species classification would be beneficial for tree selection applications.
Although we restricted our evaluations on \ac{DBH} estimation and not quality of the maps, note that the latter can play an important role in other automated tasks such as navigation and tree grasping.
Evaluating the possible use of our maps for these purposes would be of interest.
While we did not attempt to measure trees whose \ac{DBH} was less than \SI{10}{\cm}, there is interest in being able to detect and measure those small trees for regeneration monitoring purposes, as well as to avoid damaging them during operations.
More work is needed to achieve accurate measurement of those small trees from mobile lidar.
Such accuracy could possibly be achieved by combining the range measurements of lidar with angular measurements made with a camera.

\begin{acknowledgement}
  This work was supported by a Mitacs Accelerate grant and FORAC.
  We thank the Canadian Space Agency for lending us their Velodyne HDL-32, the For\^et Montmorency staff, especially Charles Villeneuve, for his help with tree measurements as well as Simon-Pierre Desch\^enes and Philippe Dandurand for their help with field work.
\end{acknowledgement}

\bibliography{references}

\end{document}